\documentclass[runningheads]{llncs}
\usepackage{graphicx}
\usepackage{gensymb}

\usepackage{tikz}
\usepackage{comment}
\usepackage{amsmath,amssymb} % define this before the line numbering.
\usepackage{color}
\usepackage{algorithm,algpseudocode}
\usepackage[T1]{fontenc}

\usepackage{algorithmicx}
\usepackage[accsupp]{axessibility}  % Improves PDF readability for those with disabilities.
\usepackage{subcaption}

\begin{document}
% \linenumbers
\pagestyle{headings}
\mainmatter
\def\ECCVSubNumber{1885}  % Insert your submission number here

% \title{Deep 6D pose estimation with Adversarial samples} % Replace with your title
\title{Adversarial samples for deep monocular 6D object pose estimation}
%\begin{comment}
% \titlerunning{ECCV-22 submission ID \ECCVSubNumber} 
% \authorrunning{ECCV-22 submission ID \ECCVSubNumber} 
% \author{Anonymous ECCV submission}
% \institute{Paper ID \ECCVSubNumber}
%\end{comment}
%******************

% \begin{comment}
\titlerunning{U6DA}
\author{Jinlai Zhang\inst{1} \and
Weiming Li\inst{2} \and
Shuang Liang\inst{2} \and
Hao Wang\inst{2} \and
Jihong Zhu\inst{1,3}}
\authorrunning{Zhang et al.}

\institute{Guangxi University \and 
Samsung Research China - Beijing (SRC-B) \and  Tsinghua University \\}
% \email{cuge1995@gmail.com, weiming.li@samsung.com, liangshuang@xs.ustb.edu.cn, whjob_20160419@sina.com, jhzhu@tsinghua.edu.cn}
% \\}
% \end{comment}
%******************
\maketitle

\begin{abstract}

Estimating 6D object pose from an RGB image is important for many real-world applications such as autonomous driving and robotic grasping. Recent deep learning models have achieved significant progress on this task but their robustness received little research attention. In this work, for the first time, we study adversarial samples that can fool deep learning models with imperceptible perturbations to input image. In particular, we propose a Unified 6D pose estimation Attack, namely U6DA, which can successfully attack several state-of-the-art (SOTA) deep learning models for 6D pose estimation. The key idea of our U6DA is to fool the models to predict wrong results for object instance localization and shape that are essential for correct 6D pose estimation. Specifically, we explore a transfer-based black-box attack to 6D pose estimation. We design the U6DA loss to guide the generation of adversarial examples, the loss aims to shift the segmentation attention map away from its original position. We show that the generated adversarial samples are not only effective for direct 6D pose estimation models, but also are able to attack two-stage models regardless of their robust RANSAC modules. Extensive experiments were conducted to demonstrate the effectiveness, transferability, and anti-defense capability of our U6DA on large-scale public benchmarks. We also introduce a new U6DA-Linemod dataset for robustness study of the 6D pose estimation task. 
% Our codes and dataset will be available online.
Our codes and dataset will be available at \url{https://github.com/cuge1995/U6DA}.
\end{abstract}

% to do 0302 1: main fig 1,2. 
% 2 sec4 title rewrite.
% 3 averaged number in table 3 or others.
% 4 grammer checking

\section{Introduction}

Understanding object pose in the three-dimensional world is an important research topic in computer vision. In the past decade, extensive research efforts have been made and achieved significant progress. In particular, recent deep learning based methods show that even using only a single RGB image without any additional depth data, 6D object pose can be predicted with high accuracy on large-scale public benchmarks~\cite{peng2019pvnet,wang2021gdr,li2019cdpn}. Such monocular 6D object pose estimation models shed light on the possibility of single-RGB sensor based solutions for various 3D applications such as augmented reality~\cite{arbarros2018fusion,aryuan2020single}, autonomous driving~\cite{sdmanhardt2019explaining,sdmanhardt2019roi} and robotic grasping~\cite{wang2019densefusion,he2021ffb6d,tian2020robust}. In these real-world scenarios, model robustness is a crucial issue to consider. In some recent research, deep learning models are found to be vulnerable to adversarial samples~\cite{dong2020benchmarking,mifgsmdong2018boosting} in 2D image classification task. The adversarial samples contain human imperceptible perturbations to the input and can result wrong model prediction. Does such problem apply to other tasks? Inspired by such concerns, to the best of our knowledge, this work for the first time studies adversarial samples for the monocular 6D object pose estimation task. To mention, our discussion is focused on instance-level 6D object pose estimation with known object 3D models. Some other works study category-level object pose estimation for unseen objects, which are beyond the scope of this paper’s discussion.

% 6D pose estimation is an important foundation for various high-level computer vision tasks, including augmented reality~\cite{arbarros2018fusion,aryuan2020single}, autonomous driving~\cite{sdmanhardt2019explaining,sdmanhardt2019roi} and robotic grasping~\cite{wang2019densefusion,he2021ffb6d,tian2020robust}. The recent explosion of 6D pose estimation algorithms are arguably a result of the employment of deep neural networks (DNNs). However, DNNs are known to be vulnerable to adversarial samples~\cite{dong2020benchmarking,mifgsmdong2018boosting}, which are human imperceptible perturbations to the input can result wrong model prediction, and have been studied extensively in image classification task. To the best our knowledge, there are exist no literature explored the adversarial sample in 6D pose estimation task before.

Following the conventions in literature, SOTA deep learning methods for 6D object pose estimation can be roughly divided into three classes: direct methods, key-point based methods and dense coordinate based methods. The latter two classes follow two-stage scheme. The direct methods ~\cite{wang2021gdr,xiang2017posecnn} directly output pose parameters from the input RGB image. With two-stage scheme, the key-point based methods~\cite{peng2019pvnet} use DNN models to detect 2D keypoints in the image first, and then solve a RANSAC-based Perspective-n-Point (PnP) problem~\cite{lepetit2009epnp} for 6D pose estimation. The dense coordinate based methods~\cite{li2019cdpn} use DNNs to build dense 2D-3D correspondences, and then estimate 6D pose by solve RANSAC-based PnP problem. Recent research results on public benchmark show that all the top-performing algorithms follow the two stage scheme. It’s noticeable that the two-stage schemes include RANSAC-based PnP that are known to be robust to noise. Meanwhile, the un-differentiable part of RANSAC-based PnP makes it impossible to implement white-box attack~\cite{goodfellow2014explaining}. These features indicate that a direct application of straight-forward adversarial attack methods may not work for these models.

% 6D pose estimation with RGB image aims to predict the 6D pose of objects within the image, which can be roughly divided into three classes, direct methods, key-point based methods and dense coordinate based methods. The latter two are followed a two stage scheme. The key-point based methods~\cite{peng2019pvnet} use DNNs to detect 2D keypoints in the image first, and then solve a RANSAC-based Perspective-n-Point (PnP) problem~\cite{lepetit2009epnp} for 6D pose estimation. The dense coordinate based methods~\cite{li2019cdpn} use DNNs to build dense 2D-3D correspondences, and then estimate 6D pose by solve RANSAC-based PnP problem. The direct methods~\cite{wang2021gdr,xiang2017posecnn} directly output pose parameters from the input RGB image. Among them, all the top-performing algorithms followed the two stage scheme. 

\begin{figure}[!tp]
\centering 
% Uncomment below line to include image
\includegraphics[width=\textwidth]{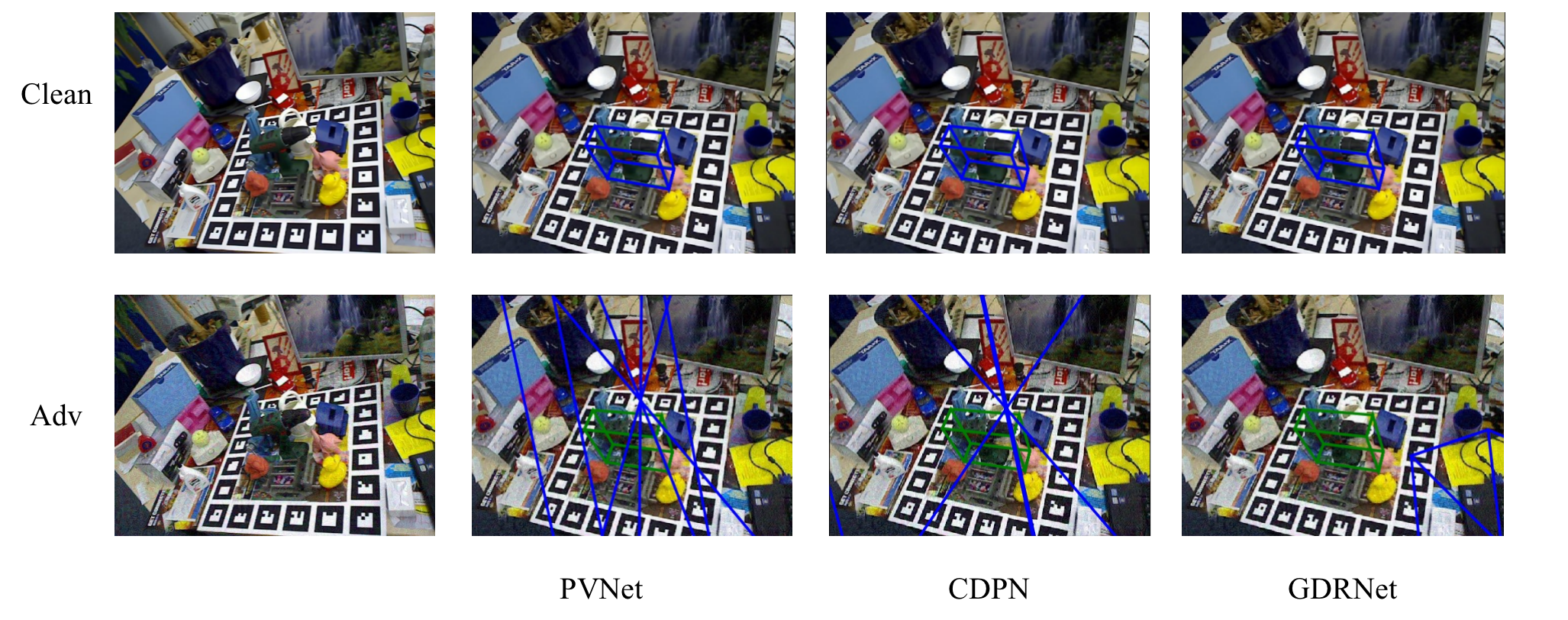}% <image name>
\caption{\label{pipeline}Comparison of pose estimation results with clean image and our U6DA generated adversarial image as inputs respectively. Column one shows a clean image and its adversarial image crafted by our U6DA. They are input to three deep learning models ~\cite{peng2019pvnet,wang2021gdr,li2019cdpn} of different methodologies, and their estimated poses for an object instance (driller in LINEMOD dataset) are shown in column 2-4. In each image, green and blue box denote ground truth and the predicted pose respectively. All the three deep models predict wrong poses with the adversarial image, which shows our U6DA attack is successful.}
\end{figure}

% However, the two stage scheme brings huge challenge to adversarial attack on 6D pose estimation task. Fisrtly, the RANSAC-based PnP~\cite{di2021so} are known to more robust to noise. Secondly, the undifferentiable part of RANSAC-based PnP making it impossible to implement white-box attack~\cite{goodfellow2014explaining}. Given the observation that three mainstream methods of 6D pose estimation are heavily rely on sementic information to localize instance, we infer that the adversarial samples craft by sementic segmentation task could be transfer to 6D pose estimation task.

Based on an observation that all the three classes of methods rely on object instance localization and shape, we infer that the adversarial samples crafted by attacking a segmentation task could be likely to transfer to 6D pose estimation task. To this end, we propose a Unified 6D pose estimation Attack, namely U6DA, which can successfully attack all the three classes of deep learning models for 6D pose estimation. Specifically, we explore a transfer-based black-box attack to 6D pose estimation, and we select the segmentation model as surrogate model. We design the U6DA loss to guide the generation of adversarial examples, the loss aims to shift the segmentation attention map away from its original position.

We perform extensive experimental evaluations on the proposed U6DA on both the Linemod \cite{hinterstoisser2012model} and Linemod Occlusion \cite{brachmann2014learning} dataset. 
Compared to two straight forward task-transfer attack methods~\cite{lu2020enhancingdrattack}~\cite{mifgsmdong2018boosting}
, we show that our U6DA is much more effective to 6D pose estimation attack.
We also show the transferability of our U6DA attack. The adversarial samples generated by our U6DA can significantly decrease the ADD metric of all three mainstream RGB-based 6D pose estimation networks. In addition, we make two extension studies. We first examine whether the SOTA physically-based rendering \cite{hodavn2020bopeccv} (PBR) training is more robust than normal training. Our experimental results show that the PBR training is not as robust as normal training. Second, we demonstrate that the RGB-D based 6D pose estimation networks are more robust to RGB adversary. We also perform experiments to test classical adversarial defense algorithms and show that they are less effective on our U6DA. Finally, we introduce a new U6DA-Linemod dataset for robustness test of 6D pose estimation task, which we hope will benefit the 6D pose estimation community. 
In summary, the major contributions of this work are as follows:

1. A novel adversarial attack method U6DA is proposed that is effective for several classes of SOTA deep learning models for 6D object pose estimation.

2. Extensive experiments are performed to examine the effectiveness, transferability, anti-defense capability of the U6DA method on large scale public benchmark datasets.

3. We construct a public-available U6DA-Linemod dataset for robustness test of 6D pose estimation models.

\section{Related work}
\subsection{Adversarial attacks}
Adversarial attack~\cite{goodfellow2014explaining} refers to the process of applying a slight perturbation to the original input of deep learning model to generate adversarial samples that fool the target model (also known as the victim model). Adversarial attack methods have been extensively studied in the image classification task~\cite{dong2020benchmarking,mifgsmdong2018boosting,wang2021gdr}. Most of them are in the white-box setting, where full-knowledge of the DNN is known by the attacker. However, white-box attacks are not practical for many real applications where the attacker only knows the output of networks. Therefore, the black-box attacks where the attacker has no prior knowledge or only partial knowledge of the model are getting more research attentions. Recently, there are also some works focus on 3D point clouds \cite{liu2022boosting} and 3D meshes \cite{zhang20213d_mesh_ttack}.
However, to the best of our knowledge, there is no adversarial attack to the 6D pose estimation task. In this paper, we make discussions to explore the black-box attack to 6D pose estimation task.

\subsection{6D pose Estimation with RGB image}
Following the conventions in literature, SOTA deep learning methods for monocular 6D object pose estimation with one RGB image can be roughly divided into three classes: direct methods, key-point based methods and dense coordinate based methods. Next, we give a brief review for each class of methods respectively.

\subsubsection{Direct methods.} These methods directly forecast the 6D object pose from input image, which treat the 6D pose estimation as a regression or classification task. Xiang et al. \cite{xiang2017posecnn} propose PoseCNN, a novel end-to-end 6D pose estimation model, which decouples 6D pose estimation task into several different components. For the translation branch, it firstly localizes the object center in the image and estimates the depth with respect to the camera. Then the translation is recovered according to the projection equation. The rotation is regressed a quaternion representation directly. Instead of regressing rotation directly, SSD-6D \cite{kehl2017ssd} treats rotation estimation as a classification problem via broken down the rotation space into classifiable ranges. Recently, researchers\cite{wang2021gdr,hu2020single} aim 
at mapping the Perspective-n-Point~\cite{lepetit2009epnp} (PnP)  algorithm using DNNs. For instance, the
 GDRNet~\cite{wang2021gdr} predicts the 6D pose via a Patch-PnP solver that is built upon DNNs. However, the DNNs are known to be vulnerable to adversarial attacks~\cite{goodfellow2014explaining}, it would be interesting to explore whether the direct methods are robust to adversarial samples.

\subsubsection{Key-point based methods.} The keypoint-based methods are generally more accurate compared to the direct methods. It usually uses DNNs to detect 2D keypoints in the image and then follow a Perspective-n-Point (PnP) solver to get the 6D pose. BB8 \cite{rad2017bb8} firstly uses DNNs to segment the target object and then builds 2D-3D correspondences via another DNN. Then the 6D pose can be estimated by a PnP solver. YOLO-6D~\cite{tekin2018real} uses the YOLO backbone to detect the 2D key-point of 3D bounding box corners, followed by a PnP solver to forecast 6D pose. \cite{oberweger2018making} predicts the 2D projections of 3D keypoints via a 2D heatmaps. PVNet~\cite{peng2019pvnet} is built upon the idea of voting-based key-point localization. It first trains DNNs to regress pixel-wise vectors pointing to the key-points, and then uses the vectors to vote for key-point locations. The PVNet yields pretty good performance under severe occlusion or truncation due to this vector-field representation. We therefore select PVNet as a victim model as a representative of key-point based method to test our attacks.

\subsubsection{Dense coordinate based methods.} The dense coordinate based methods solve the 6D pose estimation task via building dense 2D-3D correspondences, followed by a PnP solver to estimate the 6D pose. The dense 2D-3D correspondences are built via forecasting the 3D object coordinate of each target object pixel. CDPN~\cite{li2019cdpn} firstly using DNNs for dense coordinates forecasting. Pix2Pose \cite{park2019pix2pose} proposes a transformer loss to handle symmetric objects, 
% which can transform the predicted 3D coordinate of each pixel to its closest symmetric pose, 
thus improving the performance of 6D pose estimation of symmetric objects significantly compared to CDPN. In this paper, we select the CDPN as a victim model that represents the dense coordinate based methods.

\subsection{Dataset for 6D Pose Estimation}
There exist many datasets for 6D pose estimation. The Linemod\cite{hinterstoisser2012model} is the classical benchmark for instance-level 6D object pose estimation. Linemod Occlusion \cite{brachmann2014learning} is a variant of Linemod that makes up for Linemod that it lacks occlusion cases. In most papers \cite{wang2021gdr,peng2019pvnet}, Linemod Occlusion is used for testing DNNs trained on Linemod. The YCB video \cite{xiang2017posecnn} is bigger than Linemod, it consists of 21 objects collected from  YCB object set \cite{calli2015ycb}. And the various lighting conditions, severe occlusions make it much more challenging than Linemod. The T-LESS \cite{hodan2017t} aims to push the limit of 6D object pose estimation with texture-less objects. The HomebrewedDB \cite{kaskman2019homebreweddb} focus on household objects and low-textured industrial objects. The above datasets are mainly focused on occlusion and texture-less of objects. It can be seen that, compared to ImageNet variants~\cite{hendrycks2018benchmarkingimagenetc} in the field of image classification, a large-scale public dataset to evaluate the algorithm robustness is absent in the field of 6D object pose estimation.

\section{Methodology}
In this section, we introduce our U6DA algorithm. Given an RGB
image, the U6DA generates an adversarial perturbation that aims at
confusing deep 6D pose estimation networks.
\subsection{Problem Definition}
In this paper, we focus on the adversarial attacks on instance level monocular 6D pose estimation. Given an RGB image $I$, our goal is to generate an adversarial image $I_{adv}$ that can fool deep learning models of 6D pose estimation. Firstly, we give a brief introduction of deep 6D pose estimation from RGB image. The deep 6D pose estimation networks aim to estimate the object pose of the target instance when given an RGB image $I$ and the target 3D model $\mathcal{M}$. The 6D pose can be represented as rotation $\mathcal{R} \in SO(3)$ and translation $\mathcal{T} \in R^3$ w.r.t the camera. The whole task can be described as follows:
\begin{equation}
% \centering
\mathcal{[R|T]} =F\{[I,\mathcal{M}| \theta\}
\end{equation}
where  $F$ refers to a specific deep 6D pose estimation network, and $\mathcal{\theta}$ refers to the network's parameters. We aim to generate an adversarial example $I^{adv}=I+\epsilon$, which is distorted by carefully designed perturbation $\epsilon$ but will mislead the 6D pose estimation network. Typically, $\ell_p$-norm is commonly adopted to regularize the perturbation. Here, we define the generation of adversarial examples in the following.
\begin{equation}
\arg \underset{I^{adv}}{\min }\; ADD, \;
s.t.\;\left\|I-I^{adv}\right\|_p \leq \epsilon,
\label{equ:definition}
\end{equation}
where the $ADD$ is pose accuracy metric following the definition in~\cite{hodavn2016evaluation,hinterstoisser2012model} , and $p=\infty$ in this paper. Owing to two stage scheme of deep 6D pose estimation networks, it is impossible  to solve the above  problem via the famous Fast Gradient Sign Method~\cite{fgsm} (FGSM) or other white-box attacks. Instead, inspired by~\cite{ilyas2019adversarialnotbug}, we generate adversarial examples via transfer-based attack.

\subsection{Unified 6D Pose Estimation Attack}
The overall framework of our Unified 6D Pose Estimation Attack (U6DA) is shown in Algorithm \ref{u6daalgo}. The U6DA generates adversarial examples via a pretrained segmentation network. The intuition of our U6DA is simple, since the deep 6D pose estimation networks heavily rely on the semantic segmentation information of the target instance to localize the object, we infer that an attack on the segmentation network could transfer to the deep 6D pose estimation network. Therefore, we use the classical segmentation network UNet~\cite{ronneberger2015unet} as our surrogate model.

\begin{algorithm}[!t]
\caption{Unified 6D Pose Estimation Attack (U6DA) to fool deep 6D pose estimation networks.}
\label{u6daalgo}
\textbf{Input}: Original RGB image $I$, trained UNet model $\mathcal{F}_{\theta}()$, perturbation clipping factor $\epsilon$, max iteration $M$.

\textbf{Output}: Perturbed RGB image $I_{adv}$.
% \begin{spacing}{1.2}
\begin{algorithmic}[1]
\State set initial $I_{adv}^{0}=I$, $g_{0}=0$, $\alpha=\epsilon/M$ 
\State \textit{\{Calculate attention map $M_{ori}$ of $I_{adv}^{0}$\}} \Comment{Eq.~\ref{eq:seg-cam}}
\State get shfited attention map $M_{sh}$
% \State get shfited seg cam heatmap $I_{sf_cam} = T(I_{ori_cam})$
\While {$i<M$}
    \State \textit{\{Calculate attention map $M_{adv}$ of $I_{adv}^{i}$\}} \Comment{Eq.~\ref{eq:seg-cam}}
    \State $g^{i}= \frac{\partial L_{u6da}(M_{adv} - M_{sh})}{\partial I_{\mathrm{adv}}^{i}}$
    \State $g^{i+1} =g^{i}+  \frac{g^{i}}{||g^{i}||_{1}}$
    \State $I_{\mathrm{adv}}^{i+1}= \mathrm{clip}_{\epsilon}(I_{\mathrm{adv}}^{i} - \alpha * sign(g^{i+1}))$
%     %\State clip $V'$ in range $[V^0-\epsilon, V^0+\epsilon]$
%     % \State Skeleton realignment $V'=\mathrm{SSR(V')}$ 
    \State $i=i+1$ 
\EndWhile

\State \Return $I_{adv}^{i}$
\end{algorithmic}
% \end{spacing}
\end{algorithm}

To pursue high transferability of U6DA to three mainstream deep 6D pose estimation networks, we need to find common features of those networks. Inspired by~\cite{ilyas2019adversarialnotbug} and the analysis in the previous section, we firstly select deep segmentation networks to learn the general features. Moreover, we attack on the attention map to avoid overfitting in the segmentation task and fool the models to predict wrong results for object shapes that are essential for correct 6D pose estimation. Specifically, we design the U6DA loss to guide the generation of adversarial examples, the loss aims to shift the segmentation attention map away from its original position. The key idea of our U6DA is if we shift the general features away from their original position, the following RANSAC-based PnP solver would make wrong estimation. 

Next, we describe U6DA step by step. We use the Seg Grad CAM~\cite{seg_grad_cam} as the attention map due to its impressive performance in explaining segmentation networks. We therefore first introduce the Grad CAM~\cite{selvaraju2017gradcam} and Seg Grad CAM~\cite{seg_grad_cam} as follows:

\begin{figure}[!b]
\centering 
% Uncomment below line to include image
\includegraphics[width=8cm]{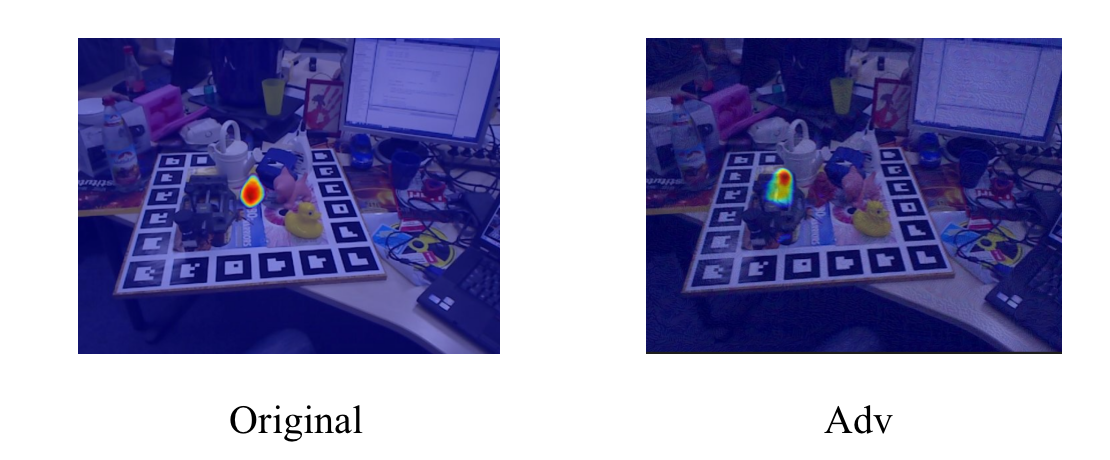}% <image name>
\caption{\label{fig1}Attention map visualization of clean image and adversarial image. The attention map of adversarial image is away from its original position.}
\end{figure}

\textbf{Grad CAM.} The Grad CAM is designed for explain the classification models, which is calculated as follows:
\begin{equation}
    \label{eq:grad-cam}
    M^c = \mathrm{ReLU}\biggl(\sum_k \alpha_c^k A^k \biggr)
    \ \ \text{with}\ \
    \alpha_c^k = \frac{1}{N} \sum_{u,v} \frac{\partial y^c}{\partial A_{uv}^k}
\end{equation}
where $\{A^k\}_{k=1}^K$ is a set of selected feature maps of interest ($K$ kernels of the last convolutional layer of a classification network), and $y^c$ is the logit for a chosen class $c$.
Grad-CAM averages the gradients of $y^c$ with respect to all $N$ pixels (indexed by $u,v$) of each feature map $A^k$ to produce a weight $\alpha_c^k$ to denote its importance.

\textbf{Seg Grad CAM.} The Seg Grad CAM is the extension of Grad CAM that can interpretation of segmentation problems. It is defined as follows:
\begin{equation}
    \label{eq:seg-cam}
    M_{seg}^c = \mathrm{ReLU}\biggl(\sum_k \alpha_c^k A^k \biggr)
    \ \ \text{with}\ \
    \alpha_c^k = \frac{1}{N} \sum_{u,v} \frac{\partial \sum_{(i,j) \in \mathcal{M}} y_{ij}^c}{\partial A_{uv}^k}
\end{equation}
where $\mathcal{M}$ is a set of pixel indices of interest in the output mask. 
% \subsection{Loss Function}

Let $M_{seg}^t$ stand for the attention map for the input $I$ and a specified class $t$. We push the attention map away from its original position. Therefore, our loss would be:
\begin{equation}
        L_\mathrm{u6da}(I) =  \|M_{adv} - M_{sh}\|_1
\end{equation}
where the $M_{adv}$ is the attention map from the adversarial sample, the $M_{sh}$ is the shifted attention map from the clean sample. 
% The basic idea of U6DA is to shift the attention map away from the original position, as illustrated in Fig. \ref{fig1}. If the attention map goes to wrong position, the key-point or dense-correspondense based method's RANSAC-based PnP solver might be wrong too. In this paper, we utilize Seg Grad CAM\cite{seg_grad_cam} to calculate the attention map $L_{seg}^t$, which is good at distinguishing the attention for the target class from the others. There exist of course many other techniques for obtaining the heat map to attack, as long as $h(x, y)$ and its gradient on $x$ could be effectively calculated. L_\mathrm{u6da}(I) = \|M_{new} - M_{old}\|_1 - \|M_{new} - M_{sh})\|_1

The adversarial samples are generated in an update process guided by the U6DA loss $L_\mathrm{u6da}$. As shown in Algorithm \ref{u6daalgo}, we set $I^0_\mathrm{adv} = I$ and the update it by momentum iterative method~\cite{mifgsmdong2018boosting}.
% procedure could be described as the following:
% \begin{equation}
% I^{k+1}_\mathrm{adv} = \text{clip}_\varepsilon\left(I^{k}_\mathrm{adv} -  \frac{g(I^{k}_\mathrm{adv})}{||g(I^{k}_\mathrm{adv})||_1} \right)
% \end{equation}
% \begin{equation}
% g(I) = \frac{\partial L_\mathrm{u6da}(I)}{\partial I}
% \end{equation}
Followed previous works~\cite{goodfellow2014explaining,mifgsmdong2018boosting}, we restrict our attack via the $\ell_\infty$ distance.

Because its attention maps have shifted away from the original position, U6DA could be used for the black-box attack to the mainstream 6D pose estimation networks. Due to the RANSAC-based PnP solver could be affected by our U6DA, the generated adversarial samples tend to be aggressive to the mainstream 6D pose estimation networks.

%  \State $g = \frac{\partial L_{u6da}(I_{adv})}{\partial I_{\mathrm{adv}}^{k}}$

\section{Experimental Results}
% In this section, we first describe the experimental settings in Sec. 4.1, then we perform the comparative study to select the parameters of U6DA in Sec. 4.2. The experiment results are present in Sec. 4.3. Moreover, to systematically study the robustness of 6D pose estimation task, we exam whether the PBR trained model more robust, whether RGB-D based deep 6D pose estimation networks robust to RGB adversary, whether the classical adversarial defense algorithm effective to U6DA. Lastly, we introduce the U6DA-Linemod dataset for robustness test of 6D pose estimation task. 

\subsection{Experimental settings}
\subsubsection{Dataset and models.}

\textbf{Linemod} \cite{hinterstoisser2012model} consists of 13 RGBD sequences, and the ground-truth 6D poses of each object are provided for each sequences. The image size is $640 \times 480$. Following previous studies~\cite{wang2021gdr,peng2019pvnet}, we select 15\% of images of each sequence for training and the remaining 85\% for testing. Note that of the RGBD images, we use only the RGB images as input.
% Each sequence includes challenging cluttered scenes, texture-less objects and light condition variations, bringing difficulties for accurate object pose prediction. Though containing sequences, the dataset is always only used for object pose detection rather than tracking, as the sequences are short and the  train/test split has made them  non-continuous.
\textbf{Linemod Occlusion} \cite{brachmann2014learning} is generated from Linemod dataset, and it was generally used for test the performance of 6D pose estimation networks under occlusion conditions. We use Linemod and Linemod Occulusion dataset. Three SOTA models are selected as the target model, which are the GDRNet~\cite{wang2021gdr}, PVNet~\cite{peng2019pvnet}, CDPN~\cite{li2019cdpn}. For segmentation networks, we use Pytorch to build the segmentation networks, such as Unet~\cite{ronneberger2015unet}, Unet$++$~\cite{zhou2019unet++}, FPN~\cite{kirillov2017unifiedfpn}, DeepLabV3~\cite{florian2017rethinking}.

\subsubsection{Metrics.} We use the ADD metric as our evaluation metric. Following the definition in~\cite{hodavn2016evaluation,hinterstoisser2012model}, if the distance is less than 10\% of the model’s diameter, we suppose the estimated pose is correct.
% ADD metric \cite{hodavn2016evaluation,hinterstoisser2012model} computes the mean distance between two transformed model points using the estimated pose and the ground-truth pose. When the distance is less than 10\% of the model’s diameter, it is claimed that the estimated pose is correct. 
% Which is defined as follows:
% \begin{equation}
% ADD=\frac{1}{m}\sum_{\mathbf{x}\in\mathcal{M}}\|(\mathbf{R}\mathbf{x}
% +\mathbf{t})-(\mathbf{\hat{R}}\mathbf{x}+\mathbf{\hat{t}})\|
% \end{equation}
% where $m$ is the number of points on the 3D CAD model, $\mathcal{M}$ is the set of all 3D points of this model, $[\mathbf{R}|\mathbf{t}]$ is the ground truth pose and 
% $[\mathbf{\hat{R}}|\mathbf{\hat{t}}]$ is the predicted pose.

% For symmetric objects, ADD metric is often replaced by ADD-S metric, where the mean distance is computed based on the closest point distance. When models trained on datasets like YCB video \cite{xiang2017posecnn} are evaluated, the ADD(-S) AUC (Area Under Curve) should also be computed as a metric.

% to do 1. the exp to GDRNet and CDPN, 2. the exp to Linemod Occlusion
% done exp, down a lot

\subsubsection{Implementation details.} We implement our U6DA attack in Pytorch~\cite{paszke2019pytorch}, the perturbation size is $\epsilon = 16$ for all experiments with pixel values in [0, 255] and the attack steps is 5. We train the segmentation models based on Pytorch using ResNet34~\cite{he2016deepresnet} as encoder, the training set is from Linemod. The epoches of training is 40, in the first 25 epoches, the learning rate is 0.001, and the later epochs are with a learning rate 0.00001.

\subsection{Sensitivity analysis}
Here, we conduct a series of experiments on Linemod dataset to analyze the following aspects of our U6DA. All the attacks are implemented on benchvise instance to PVNet.

% fgsm
% 2d projections metric: 0.3236434108527132
% ADD metric: 0.22189922480620156
% 5 cm 5 degree metric: 0.17635658914728683
% mask ap70: 0.3953488372093023

% ifgsm
% 2d projections metric: 0.9922480620155039
% ADD metric: 0.9651162790697675
% 5 cm 5 degree metric: 0.9825581395348837
% mask ap70: 1.0
%cam
% 2d projections metric: 0.9796511627906976
% ADD metric: 0.8982558139534884
% 5 cm 5 degree metric: 0.9302325581395349
% mask ap70: 0.998062015503876
% dist_cam
% 2d projections metric: 0.9786821705426356
% ADD metric: 0.9089147286821705
% 5 cm 5 degree metric: 0.936046511627907
% mask ap70: 1.0

% \subsubsection{The effect of loss function}
% l1 l2 mae rmse
\subsubsection{Shift Position.}
% row col or both
\begin{figure*}[!t]
    \centering
    \begin{subfigure}[t]{0.3\textwidth}
           \centering
           \includegraphics[width=\textwidth]{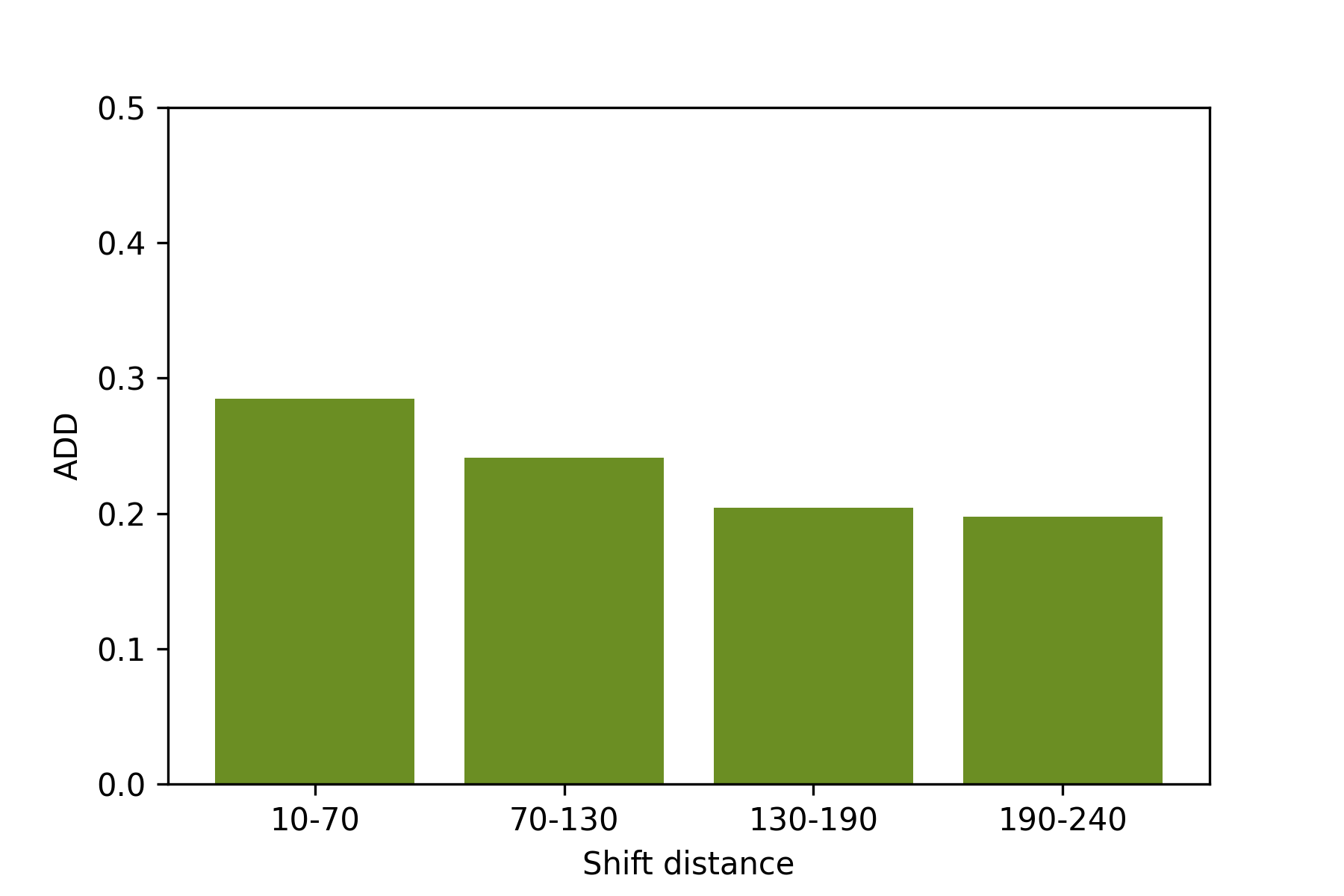}
            \caption{R test}
            \label{fig:a}
    \end{subfigure}
    \begin{subfigure}[t]{0.3\textwidth}
            \centering
            \includegraphics[width=\textwidth]{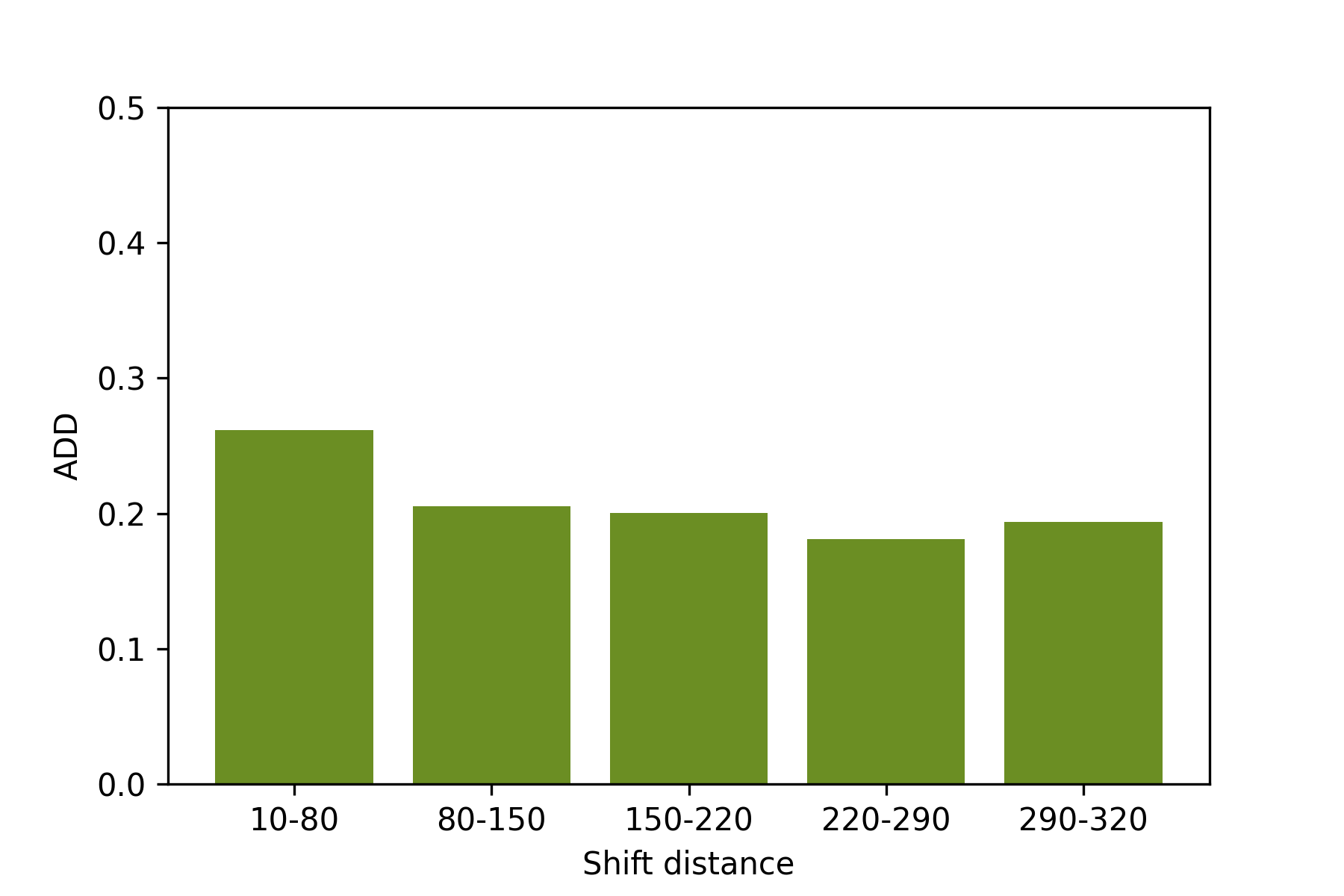}
            \caption{C test}
            \label{fig:b}
    \end{subfigure}
    \begin{subfigure}[t]{0.3\textwidth}
            \centering
            \includegraphics[width=\textwidth]{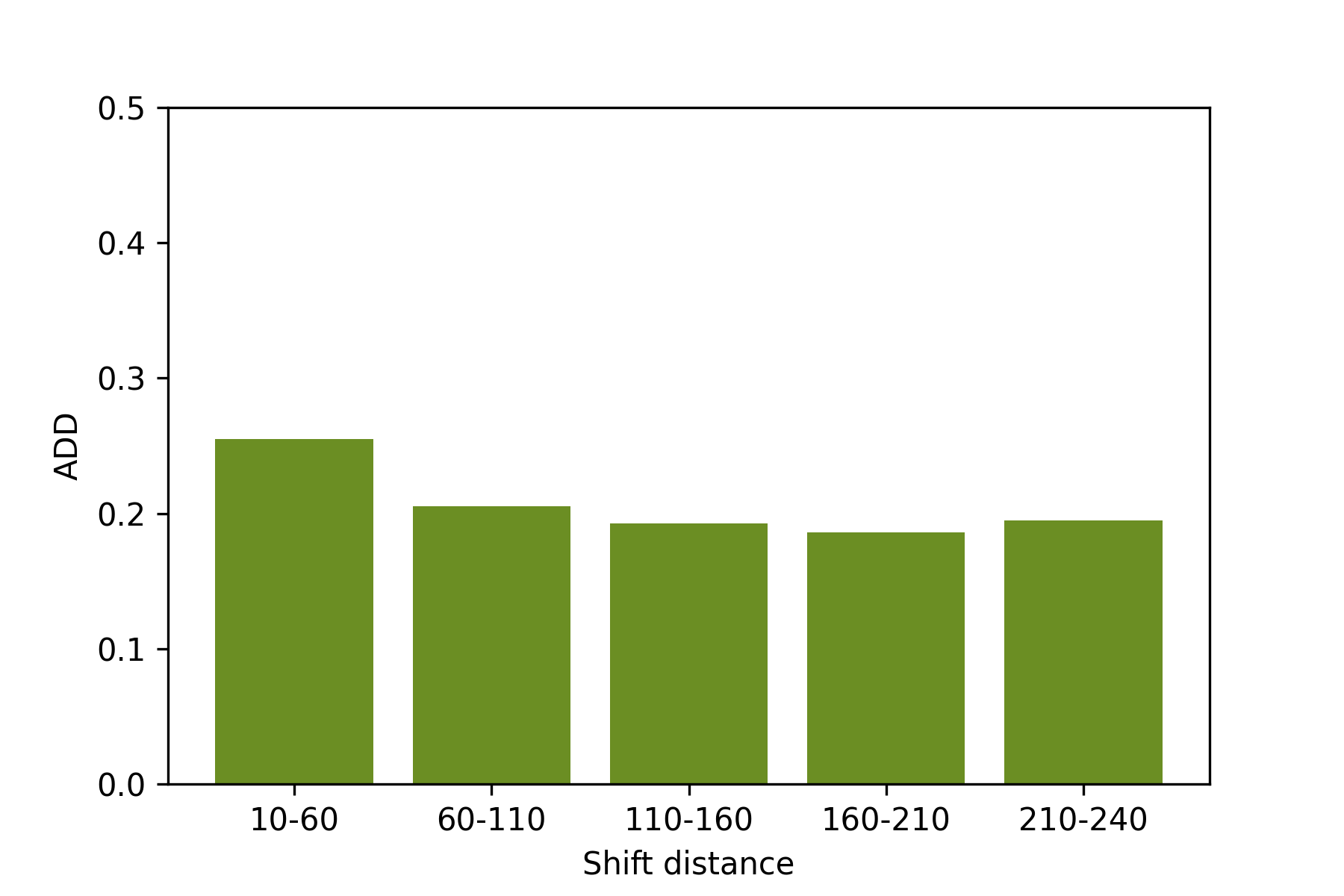}
            \caption{RC test}
            \label{fig:c}
    \end{subfigure}
    \caption{The effect of different shift distance.}
\end{figure*}
In our U6DA attack, we shift the attention map away from its original position and this leads to error in the RANSAC-based PnP solver. In this section, we explore the effect of different shift position on U6DA. In Linemode dataset, the image size is 640 pixels in width and 480 pixels in height, and most objects are located at image center. We therefore firstly perform a shift in the up and down direction of height with the maximum shift distance as half of the height (R test), and then perform shift in the left and right direction of width with the maximum shift distance as half of the width (C test). Lastly, we perform shift in both width and height (RC test). For R test, we first randomly select the up or down direction, and the shift distance is randomly generated within a certain range. We perform four different sets of experiments with shift distance 10-70, 70-130, 130-190, 190-240. For R test, we first randomly select the left or right direction, then the shift distance is randomly selected from 10-80, 80-150, 150-220, 220-290, 290-320. For RC test, both shift directions are generated followed R test and C test, the shift distance is randomly selected from 10-60, 60-110, 110-160, 160-210, 210-240. The experimental results are shown in Figure 3 a,b,c. It can be seen that the width shift is more effective for U6DA. In order to avoid overfitting to PVNet, we select width and height shift with 110-160 shift distance in the following experiments.

\subsubsection{The effect of different segmentation models.} Since our U6DA attack is firstly performed on the segmentation model and then transfer to the 6D pose estimation models, in this section we explore the effect of different segmentation models on the transferability. We perform U6DA attack on the Unet~\cite{ronneberger2015unet}, Unet++~\cite{zhou2019unet++}, FPN~\cite{kirillov2017unifiedfpn}, DeepLabV3~\cite{florian2017rethinking} and their ensembles. The results are shown in Table \ref{com_seg}, the ensemble attack might increase the attack performance but requires much more computing resources. Unlike the adversarial attack of image classification that ensemble victim model will improve the transferability considerably \cite{mifgsmdong2018boosting}, our ensembled models do not always bring a performance improvement. For example, the combination of all models achieves relatively poor performance. Moreover, although the DeepLabV3 achieved the best attack performance over single model, it was slower than Unet, the DeepLabV3 needs 3.24s in every iteration of our U6DA attack while Unet only needs 0.08s in NVIDIA RTX 3090 GPU, we therefore select the Unet as our surrogate model in the following experiments.

\begin{table}[tp!]
\tabcolsep=0.17cm
\begin{center}
\caption{The ADD of our U6DA attack based on different segmentation models. The clean ADD is 99.52.}
\label{com_seg}
% \scalebox{0.8}{
\begin{tabular}{c|c}
\hline
Victim models & ADD\\
\hline
UNet & 19.65 \\
UNet++ & 46.12 \\
FPN & 28.19 \\
DeepLabV3 & \bf16.27 \\
\hline
UNet + UNet++ & 20.63\\
UNet + FPN & 25.48\\
UNet + DeepLabV3 & 23.93\\
FPN + UNet++ & 22.48\\
DeepLabV3 + UNet++ & \bf19.57\\
FPN + DeepLabV3 & 26.06\\
\hline
UNet + FPN + UNet++ & \bf14.92\\
UNet + DeepLabV3 + UNet++ & 36.82\\
UNet + FPN + DeepLabV3& 21.70\\
FPN + DeepLabV3 + UNet++ & 27.13\\
\hline
UNet + FPN + DeepLabV3 + UNet++ & 40.69\\
\hline
\end{tabular}                  
% } % scalebox
\vspace{-0.2mm}
\end{center}
\vspace{-2mm}

\end{table}

\subsubsection{The effect of different perturbation size.} 
\begin{figure*}[!t]
    \centering
    \begin{subfigure}[t]{0.23\textwidth}
           \centering
           \includegraphics[width=\textwidth]{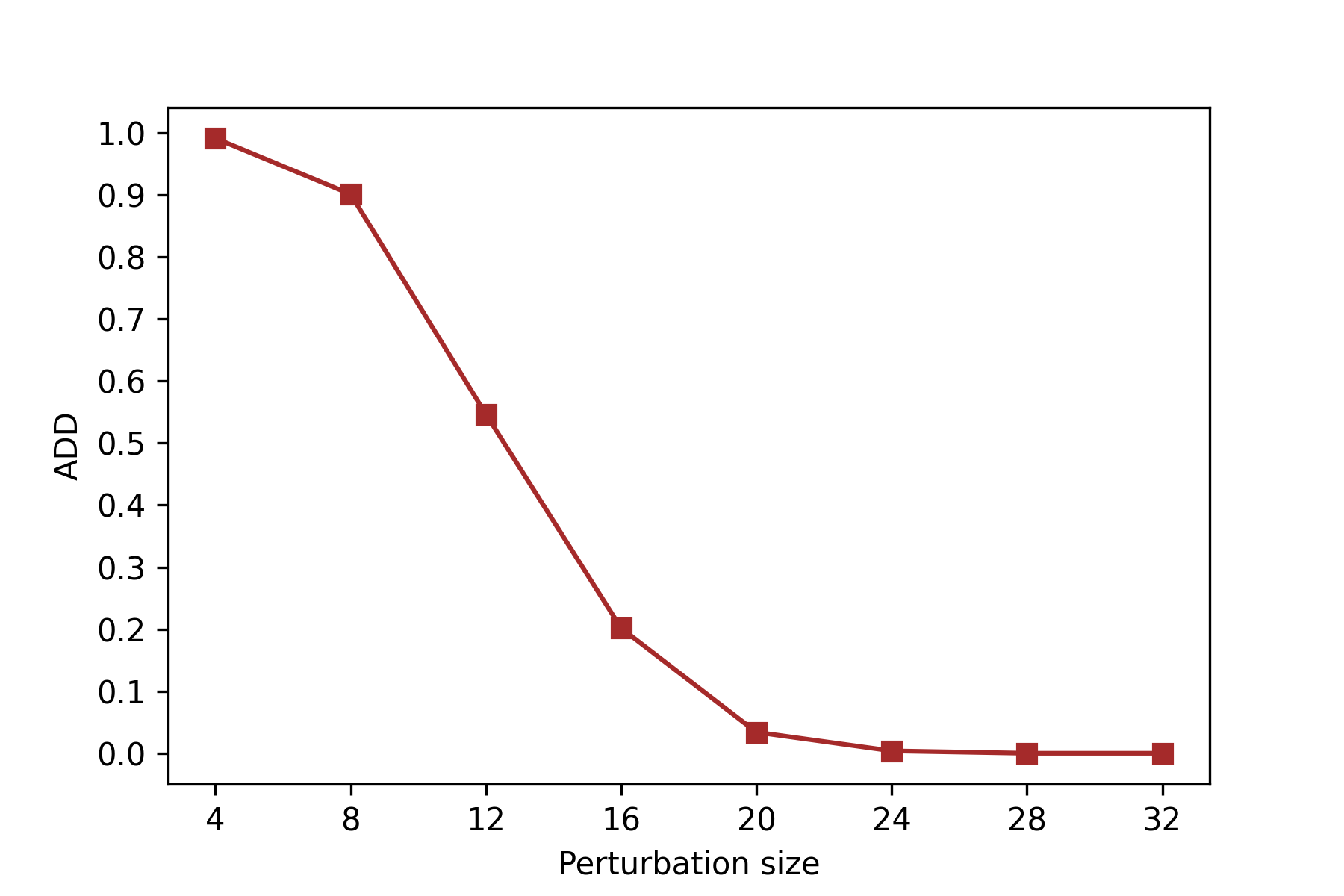}
            \caption{ADD}
            \label{ps:a}
    \end{subfigure}
    \begin{subfigure}[t]{0.23\textwidth}
            \centering
            \includegraphics[width=\textwidth]{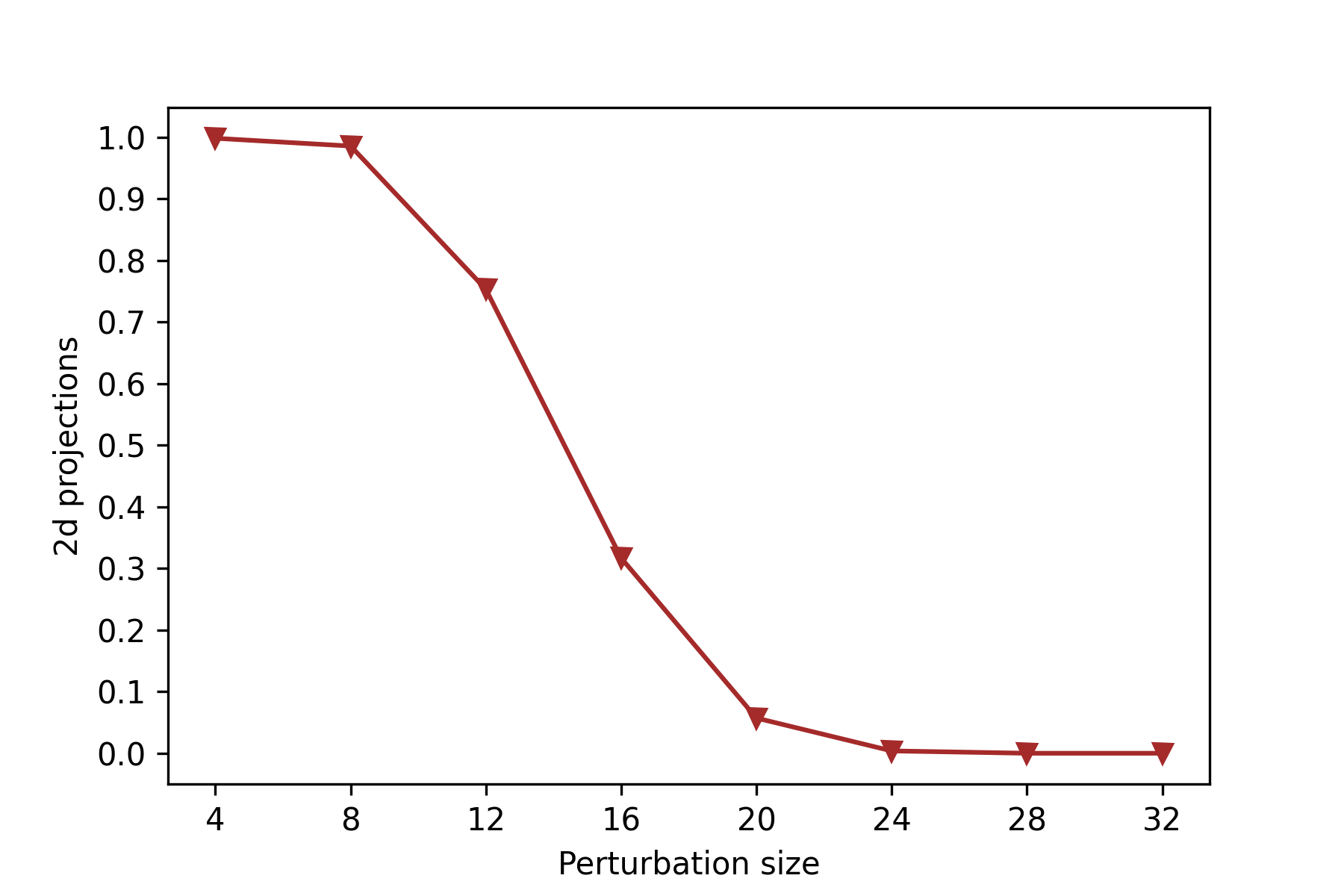}
            \caption{2d projections}
            \label{ps:b}
    \end{subfigure}
    \begin{subfigure}[t]{0.23\textwidth}
            \centering
            \includegraphics[width=\textwidth]{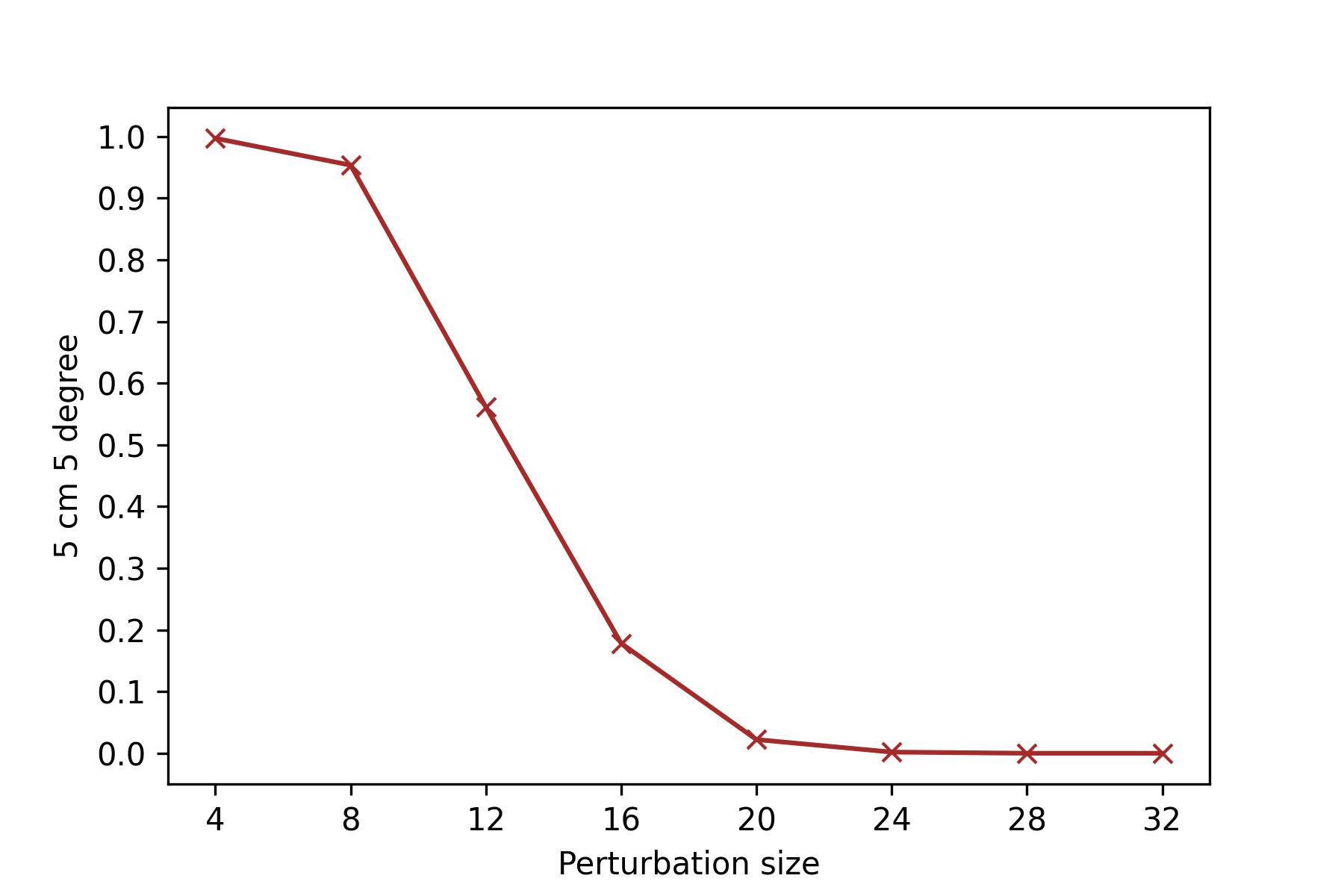}
            \caption{5$\degree$ cm 5}
            \label{ps:c}
    \end{subfigure}
    \begin{subfigure}[t]{0.23\textwidth}
            \centering
            \includegraphics[width=\textwidth]{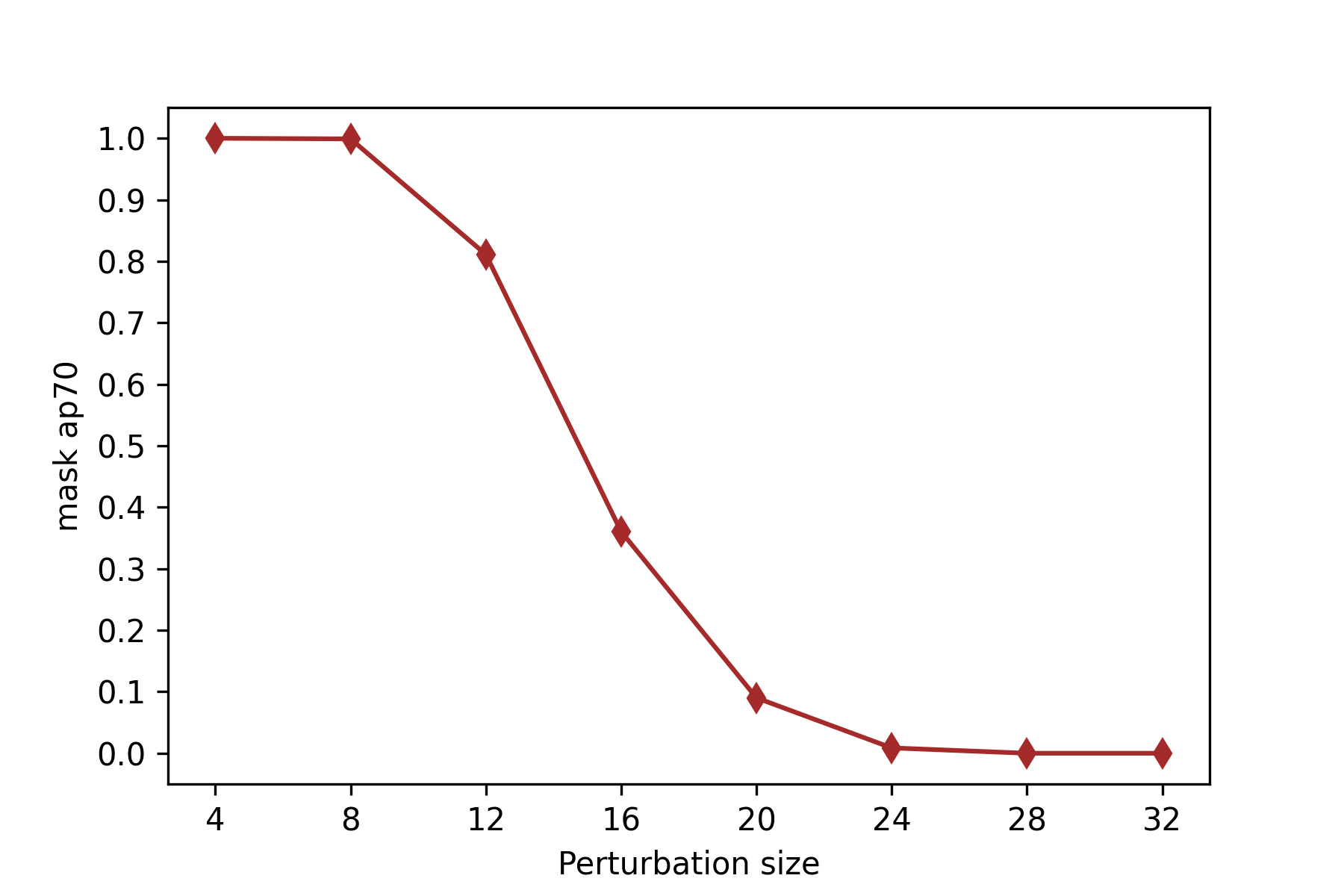}
            \caption{mask ap70}
            \label{ps:d}
    \end{subfigure}
    \caption{The effect of different perturbation size.}
    \label{ps}
    
\end{figure*}

In the adversarial attack on image domain, the $\epsilon = 16$ is a commonly used attack strength that generally considered imperceptible to humans~\cite{dong2020benchmarking,mifgsmdong2018boosting,ilyas2019adversarialnotbug,wang2020unifiedtransfer}. To gain a deep understanding of U6DA attack, we also performed attacks with $\epsilon$ in [4,8,12,16,20,24,28,32]. As shown in Figure \ref{ps}, when $\epsilon > 20$, the ADD decreased to almost zero. We also present the 2D Projection metric \cite{brachmann2016uncertainty}, 5$\degree$ 5 cm \cite{shotton2013scene} and mask ap70 \cite{peng2019pvnet}, those metrics are droped sharply from 4 to 16, and then reach almost 0 when $\epsilon > 20$.

% 

% The configuration “A+B” is the final configuration for our approach, which is denoted by “U6DA” in the following experiments.
% 1. 4
% 2d projections metric: 0.998062015503876
% ADD metric: 0.9912790697674418
% 5 cm 5 degree metric: 0.997093023255814
% mask ap70: 1.0
% 2. 8
% 2d projections metric: 0.9854651162790697
% ADD metric: 0.9011627906976745
% 5 cm 5 degree metric: 0.9534883720930233
% mask ap70: 0.999031007751938
% 3. 12
% 2d projections metric: 0.752906976744186
% ADD metric: 0.5465116279069767
% 5 cm 5 degree metric: 0.561046511627907
% mask ap70: 0.811046511627907
% 4. 16
% 2d projections metric: 0.3168604651162791
% ADD metric: 0.20155038759689922
% 5 cm 5 degree metric: 0.17829457364341086
% mask ap70: 0.36046511627906974
% 5. 20
% 2d projections metric: 0.05717054263565891
% ADD metric: 0.03391472868217054
% 5 cm 5 degree metric: 0.022286821705426358
% mask ap70: 0.09011627906976744
% 6. 24
% 2d projections metric: 0.003875968992248062
% ADD metric: 0.003875968992248062
% 5 cm 5 degree metric: 0.001937984496124031
% mask ap70: 0.00872093023255814
% 7. 28
% 2d projections metric: 0.0
% ADD metric: 0.0
% 5 cm 5 degree metric: 0.0
% mask ap70: 0.0
% 8. 32
% 2d projections metric: 0.0
% ADD metric: 0.0
% 5 cm 5 degree metric: 0.0
% mask ap70: 0.0

% \subsection{Experiment on U6DA}

% \subsubsection{The effect of different steps}

% \subsubsection{The effect of different $l_{\infty}$}

% \subsubsection{The effect on segmentation}
% 1. iou

% 2. visual comparision

\subsection{Comparasion with SOTA cross task attack.} 
Our U6DA transfer the adversarial examples from segmentation models to the 6D pose estimation models, which is a cross task attack. We therefore compare our U6DA with SOTA cross task attack, the DR attack~\cite{lu2020enhancingdrattack}, which aims to reduce the standard deviation of internal feature map. We also compare a strong transferable attack, MI-FGSM~\cite{mifgsmdong2018boosting}. We perform DR and MI-FGSM attack on the Unet, which is the same as our U6DA.

\begin{table}[!t]
\tabcolsep=0.17cm
\begin{center}
\caption{The ADD of our U6DA compared to the SOTA corss task attack. The clean denotes the clean data.}
\label{com_sota}
% \scalebox{0.8}{
\begin{tabular}{c|c|c|c|c}
\hline
Victim model & clean &U6DA attack & DR attack & MI-FGSM\\
\hline
PVNet & 99.52 &\bf19.65 & 91.18 & 51.93\\
\hline
\end{tabular}                  
% } % scalebox
\vspace{-0.2mm}
\end{center}
\vspace{-2mm}

\end{table}
As shown in Table \ref{com_sota}, our U6DA outperform DR and MI-FGSM attack significantly when transfered to PVNet. We infer that due to the 6D pose estimation based on DNNs having a different paradigm from the classification, segmentation and object detection task, the DR attack can have high transferability between the later tasks, but have relatively lower transferability on the 6D pose estimation task.
% There are also some mothods \cite{chen2020universalaoa} that aim to distract the attention region and force the model to focus on non-target regions. However, empirically, we find those methods are uneffectively to 6D pose estimation task. The experimental results are shown in the appendix.

\begin{table}[!tp]
\tabcolsep=0.17cm
\begin{center}
\caption{The ADD of clean data and adversarial samples of three mainstream RGB based deep 6D pose estimation networks on the \textbf{LINEMOD} dataset. The clean denotes the clean data, the adv denotes the adversarial attack generated by our U6DA.}
% \scalebox{0.8}{
\label{mainresults}
\begin{tabular}{|c|cc|cc|cc|}
\hline
 Models & \multicolumn{2}{c|}{GDRNet} & \multicolumn{2}{c|}{CDPN} & \multicolumn{2}{c|}{PVNet}\\ 
\hline
          
data & clean & adv & clean & adv & clean & adv\\\hline
ape & 76.29 & 0.00 & 64.67 & 0.13 & 47.71 & 4.09 \\ 
benchwise & 97.96 & 3.98 & 96.80 & 37.21 & 99.52 & 19.65 \\
cam & 95.29 & 0.00 & 90.69 & 3.93 & 84.21 & 0.98 \\ 
can & 98.03 & 0.00 & 94.78 & 14.54 & 95.86 & 35.62 \\ 
cat & 93.21 & 0.00 & 85.53 & 2.91 & 77.64 & 17.76 \\ 
driller & 97.72 & 0.00 & 92.37 & 0.13 & 96.53 & 0.19 \\ 
duck & 80.28 & 0.00 & 66.76 & 0.64 & 55.96 & 10.79 \\ 
eggbox & 99.53 & 0.00 & 99.53 & 5.63 & 100 & 31.45 \\ 
glue & 98.94 & 0.00 & 98.75 & 6.51 & 81.27 & 17.86 \\ 
holepuncher & 91.15 & 0.00 & 87.44 & 4.24 & 81.16 & 24.17 \\ 
iron & 98.06 & 0.51 & 95.61 & 34.72 & 98.47 & 41.47 \\ 
lamp & 99.14 & 0.00 & 96.26 & 19.01 & 99.42 & 10.75 \\ 
phone & 92.35 & 0.00 & 85.84 & 13.99 & 91.74 & 27.08 \\ \hline
average & 93.69 & 0.35 & 88.85 & 11.05 & 85.34 & 18.60 \\ \hline
\end{tabular}
% } % scalebox
\vspace{-0.2mm}
\end{center}

\vspace{-2mm}

\end{table}

\subsection{Transferability Results}
In this section, we perform U6DA to three mainstream RGB based deep 6D pose estimation networks, which are the direct methods GDRNet~\cite{wang2021gdr}, the key-point based methods
PVNet~\cite{peng2019pvnet}, the dense coordinate based methods CDPN~\cite{li2019cdpn}. The results of U6DA are shown in Table \ref{mainresults}. The direct methods GDRNet are the most vulnerable to U6DA attack among three models. We infer that might be caused by the Linear Behavior~\cite{goodfellow2014explaining} of deep networks, the direct methods directly output the 6D pose from an input RGB image, a small perturbation to the input is enough to make the prediction result very biased. As for the dense coordinate based methods CDPN, which was not as robust as the key-point based methods PVNet. The ADD of all instance and networks are decreased significantly, which validate the effectiveness of our U6DA attack.

\begin{figure}[!t]
\centering 
% Uncomment below line to include image
\includegraphics[width=\textwidth]{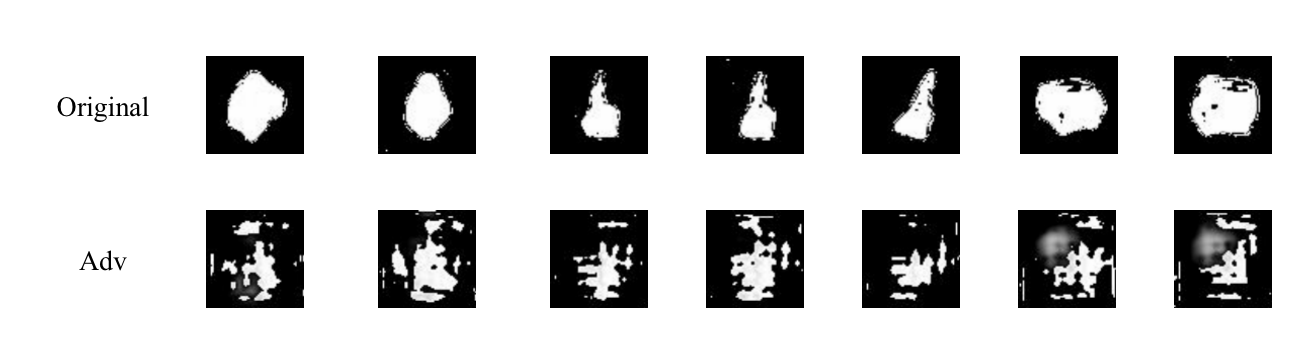}% <image name>
\caption{\label{fig_gdr}Visualization of the Visible Object Mask of GDRNet.}
\vspace{-0.0mm}
\end{figure}

\begin{figure}[!t]
\centering 
% Uncomment below line to include image
\includegraphics[width=0.75\textwidth]{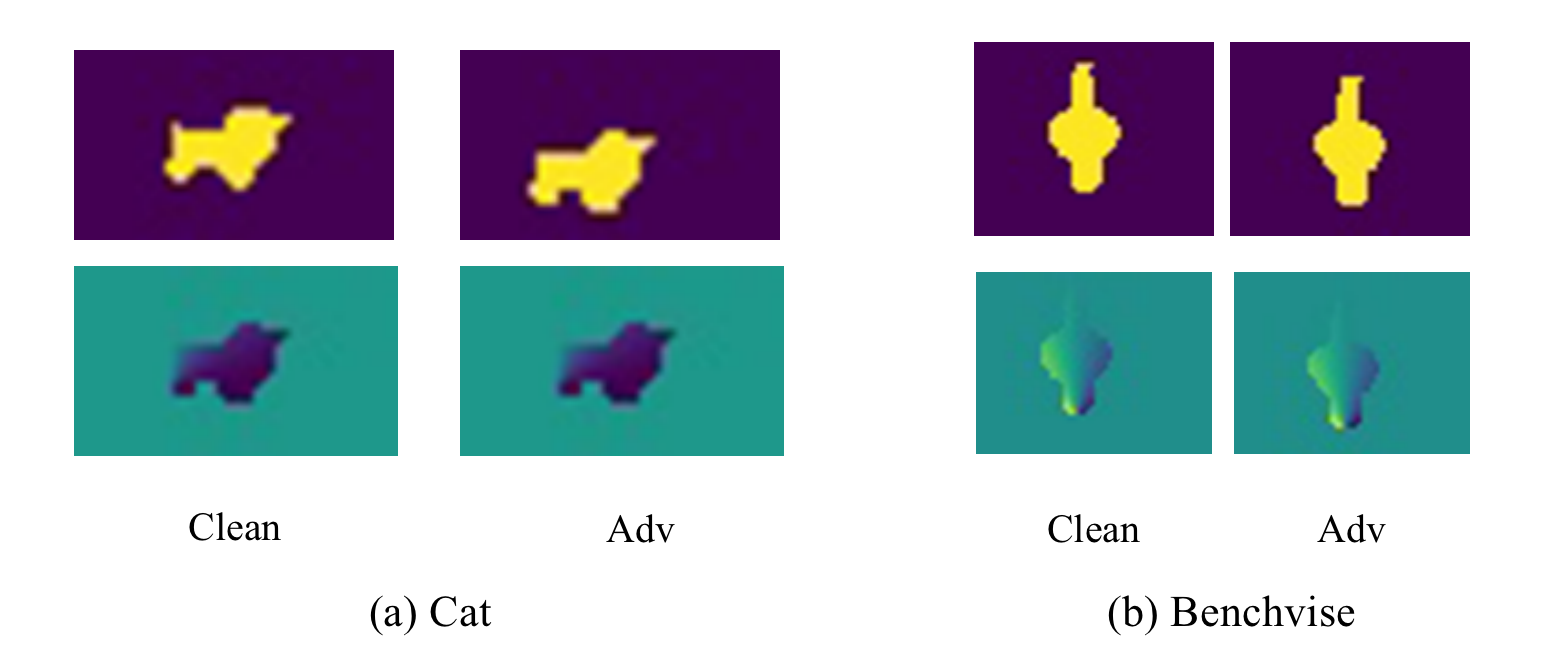}% <image name>
\caption{\label{fig_pvnet}Visualization of the vector field map (bottom row) and segmentation mask (top row) of PVNet.}
\vspace{-0.0mm}
\end{figure}

We also explore how U6DA works for PVNet and GDRNet. We visualized the vector field map and segmentation mask of PVNet with both clean sample and adversarial sample. The benchvise and cat instance are presented. For GDRNet, we randomly select some Visible Object Mask of clean samples and adversarial samples craft by our U6DA. From Figure \ref{fig_gdr} and Figure \ref{fig_pvnet}, we can observed that the Visible Object Mask of adversarial sample is relatively messy compared to the clean sample, the intermediate representation of PVNet has moved and slightly deformed after U6DA attack. Overall, the intermediate representation of PVNet is much robust than GDRNet, this might be the reason why PVNet is more robust than GDRNet under our U6DA.

% \begin{table}
% \begin{center}

% \label{mainresults}
% \begin{tabular}{|c|cc|cc|cc|}
% \hline\noalign{\smallskip}
%  Models & \multicolumn{2}{c|}{GDRNet} & \multicolumn{2}{c|}{CDPN} & \multicolumn{2}{c|}{PVNet}\\ 
% \noalign{\smallskip}
% \hline
% \noalign{\smallskip}
% data & clean & adv & clean & adv & clean & adv\\\hline
% ape & 76.29 & 0.00 & 64.67 & 0.13 & 47.71 & 4.09 \\ 
% benchwise & 97.96 & 3.98 & 96.80 & 37.21 & 99.52 & 19.65 \\
% cam & 95.29 & 0.00 & 90.69 & 3.93 & 84.21 & 0.98 \\ 
% can & 98.03 & 0.00 & 94.78 & 14.54 & 95.86 & 35.62 \\ 
% cat & 93.21 & 0.00 & 85.53 & 2.91 & 77.64 & 17.76 \\ 
% driller & 97.72 & 0.00 & 92.37 & 0.13 & 96.53 & 0.19 \\ 
% duck & 80.28 & 0.00 & 66.76 & 0.64 & 55.96 & 10.79 \\ 
% eggbox & 99.53 & 0.00 & 99.53 & 5.63 & 100 & 31.45 \\ 
% glue & 98.94 & 0.00 & 98.75 & 6.51 & 2.11 & 17.86 \\ 
% holepuncher & 91.15 & 0.00 & 87.44 & 4.24 & 81.16 & 24.17 \\ 
% iron & 98.06 & 0.51 & 95.61 & 34.72 & 98.47 & 41.47 \\ 
% lamp & 99.14 & 0.00 & 96.26 & 19.01 & 99.42 & 10.75 \\ 
% phone & 92.35 & 0.00 & 85.84 & 13.99 & 91.74 & 27.08 \\ 
% \hline
% \end{tabular}
% \end{center}
% \end{table}
% \setlength{\tabcolsep}{1.4pt}

\subsection{U6DA under defense}
We also evaluate our U6DA attack under several adversarial defenses, such as JPEG compression \cite{shin2017jpegdefense} and pixel deflection \cite{prakash2018deflectingpddefense} (PD). The results are shown in Table \ref{defense}. The PD defense slightly increased the ADD, but the JPEG compression defense decreased the ADD significantly. The results indicated that the traditional denoising-based defense is ineffective to our U6DA attack.

\begin{table}[!t]
\begin{center}
\caption{The ADD of our U6DA attack under adversarial defenses. The instance is benchvise of Linemod datset.}
\label{defense}
% \scalebox{0.8}{
\begin{tabular}{c|c|c|c}
\hline
Victim model & U6DA & JPEG & PD\\
\hline
PVNet & 19.65 &1.74 & 20.25 \\
\hline
\end{tabular}                  
% } % scalebox
% \vspace{-0.2mm}
\end{center}
\vspace{-0.0mm}

\end{table}

\subsection{Extension studies}
\subsubsection{Are PBR trained model more Robust?}

The physically-based
rendering\cite{hodavn2020bopeccv} (PBR) training are dominated in RGB-based 6D pose estimation to improve the accuracy and robustness, one intriguing question would be: are PBR trained model robust to adversarial attacks? 
To answer this question, we performed U6DA attack on GDRNet\cite{wang2021gdr} with two $\epsilon$ and two versions of GDRNet: One trained with real data from Linemod and synthetic images followed\cite{peng2019pvnet}, another one is trained with real data and PBR data. 
As shown in Table~\ref{pbr}, the answer is counterintuitive, the PBR doesn't improve the robustness to adversarial attack by our U6DA, instead, the robustness of real+synthetic training is consistently and significantly outperform the real+pbr training.

\begin{table}[t!]
\begin{center}
\caption{The ADD of clean data and U6DA attack of GDRNet~\cite{wang2021gdr} on the \textbf{LINEMOD Occlusion} dataset. Where adv16 denotes adversarial samples generated with $\epsilon = 16$ and adv8 denotes adversarial samples generated with $\epsilon = 8$. }
\label{pbr}
% \scalebox{0.8}{
\begin{tabular}{|c|ccc|ccc|}
\hline
 Models & \multicolumn{3}{c|}{Real+Syn} & \multicolumn{3}{c|}{Real+Pbr} \\
\hline
          
data & clean & adv16 & adv8 & clean & adv16 & adv8\\\hline
ape & 41.28 & 1.03 & 16.67 & 44.87 & 0.09 & 6.32 \\ 
can & 71.09 & 17.98 & 50.54 & 79.70 & 1.41 & 24.94 \\ 
cat & 18.20 & 0.08 & 1.35 & 30.58 & 0.08 & 0.51 \\ 
driller & 54.61 & 0.16 & 5.27 & 67.79  & 0.49 & 0.82 \\ 
duck & 41.64 & 20.03 & 26.95 & 39.98  & 4.72 & 17.24 \\ 
eggbox & 40.22 & 3.93 & 18.53 & 49.87 & 14.86 & 15.80 \\ 
glue & 59.49 & 27.19 & 50.28 & 73.70 & 20.31 & 48.72 \\ 
holepuncher & 52.56 & 6.12 & 29.59 & 62.73 & 1.57 & 23.88 \\ 
\hline
average & 47.39 & 9.57 & 24.89 & 56.15 & 5.44 & 17.28 \\ \hline
\end{tabular}                  
% } % scalebox
\vspace{-0.2mm}
\end{center}

\vspace{-2mm}

\end{table}

\subsubsection{Are RGB-D based deep 6D pose estimation networks robust to RGB adversary?}
The RGB-D based methods are dominated in the BOP benchmark~\cite{hodan2018bop}. One interesting question would be are RGB-D based deep 6D pose estimation networks robust to RGB adversary? To answer this, we validate in Linemod dataset. 
We perform U6DA attack on FFB6D~\cite{he2021ffb6d}, PVN3D~\cite{he2020pvn3d} and DenseFusion~\cite{wang2019densefusion}. The results are shown in Table~\ref{rgbdlinemod_add}, we can observe that the RGB-D based deep 6D pose estimation networks are more robust to RGB adversary compared to the RGB based deep 6D pose estimation networks, the ADD of RGB-D models only slightly affected by RGB adversary.

\begin{table}[!t]
\tabcolsep=0.17cm
\begin{center}
\caption{The ADD of clean data and U6DA attack of three famous RGB-D based deep 6D pose estimation networks on the \textbf{LINEMOD} dataset. The clean denotes the clean data, the adv denotes the adversarial attack generated by our U6DA attack.}
% \scalebox{0.8}{
\label{rgbdlinemod_add}
\begin{tabular}{|c|cc|cc|cc|}
\hline
 Models & \multicolumn{2}{c|}{FFB6D} & \multicolumn{2}{c|}{PVN3D} & \multicolumn{2}{c|}{DenseFusion}\\ 
\hline
          
data & clean & adv & clean & adv & clean & adv\\\hline
ape & 98.76 & 51.71 & 97.23 & 95.61 & 93.42 & 92.37 \\ 
benchwise & 100.0 & 100.0 & 99.61 & 99.41 & 93.89 & 93.79 \\
cam & 100.0 & 70.88 & 99.60 & 84.11 & 97.35 & 95.98 \\ 
can & 99.90 & 99.50 & 99.40 & 97.54 & 94.19 & 93.79 \\ 
cat & 99.90 & 90.31 & 99.80 & 93.81 & 96.91 & 96.91 \\ 
driller & 100.0 & 90.78 & 99.30 & 49.85 & 87.41 & 84.84 \\ 
duck & 98.87 & 98.12 & 98.12 & 96.71 & 92.86 & 93.15 \\ 
eggbox & 100.0 & 95.86 & 99.81 & 98.21 & 99.81 & 99.81 \\ 
glue & 99.90 & 96.23 & 99.61 & 99.03 & 99.81 & 99.90 \\ 
holepuncher & 100.0 & 74.21 & 99.90 & 98.47 & 93.24 & 91.53 \\ 
iron & 100.0 & 99.89 & 99.69 & 99.28 & 97.96 & 98.06 \\ 
lamp & 100.0 & 95.49 & 99.81 & 98.08 & 96.55 & 96.45 \\ 
phone & 99.80 & 99.90 & 99.32 & 99.51 & 96.06 & 95.67 \\ \hline
average & 99.78 & 89.45 & 99.32 & 93.05 & 95.34 & 94.79 \\ \hline
\end{tabular}
% } % scalebox
\vspace{-0.2mm}
\end{center}
\vspace{-1mm}

\end{table}

%jpeg 50
% 2d projections metric: 0.0377906976744186
% ADD metric: 0.01744186046511628
% 5 cm 5 degree metric: 0.00872093023255814
% mask ap70: 0.06395348837209303
%pd
% 2d projections metric: 0.31976744186046513
% ADD metric: 0.20251937984496124
% 5 cm 5 degree metric: 0.17829457364341086
% mask ap70: 0.3594961240310077

% m_abo_u6dappt_r110_c160_benchvise_16
% \subsection{Discussion}

\subsection{U6DA-Linemod dataset}
% In the field of 6D pose estimation, there are data sets of linemod occlusion and transparent datasets to evaluate algorithms under occlusion and transparent environment. However, there is no data set specifically used to evaluate the robustness of the network. 
% Therefore, due to our U6DA exhibits strong transferability, we believe that adversarial data is beneficial to 6D pose estimation area. 
In this section, we introduce U6DA-Linemod dataset. Following the BOP benchmark \cite{hodavn2020bopeccv}, we use U6DA to attack Unet to generate adversarial samples from all the Linemod test set, we name this dataset as U6DA-Linemod. This dataset could be used for evaluating the robustness of SOTA 6D pose estimation models, and the adversarial defenses for 6D pose estimation also could benefit from it. The U6DA-Linemod can be download online, and by replacing the original BOP Linemod test set, the robustness of SOTA 6D pose estimation models could be evaluated.

\section{Conclusion}
In this paper, we propose U6DA, the first adversarial attack to deep 6D pose estimation networks. Our U6DA selects a segmentation model as surrogate model and then crafts adversarial samples by shifting the attention map. Extensive experiments reveal that the U6DA method is effective to fool all the three mainstream classes of 6D pose estimation networks with high transferability. Various experiments are performed to show properties of the U6DA such as parameter selection and effectiveness against data augmentation. Further more, we introduce a new U6DA-Linemod dataset, the first dataset on 6D pose estimation task to evaluate the robustness and defenses. 
In the future, we plan to explore how to systematically improve pose estimation model's robustness against adversarial samples such as those generated by the U6DA.

\clearpage
\bibliographystyle{splncs04}
\bibliography{egbib}
\end{document}